\documentclass[10pt,twocolumn,letterpaper]{article}

\usepackage[pagenumbers]{paper}

\usepackage{graphicx}
\usepackage{booktabs}
\usepackage{multirow}
\usepackage{amsmath}
\usepackage{float}

\definecolor{linkblue}{rgb}{0.21,0.49,0.74}
\usepackage[pagebackref,breaklinks,colorlinks,allcolors=linkblue]{hyperref}

\title{DRIFT: Detachable Reasoning for Runtime-Flexible and Efficient \\End-to-End Driving}

\author{
Emad Aghajanzadeh$^{1,2,*}$, Bethany Anne Wong$^{1,2,*\dagger}$, Leonhard Kunze$^{1,2,\dagger}$, Moritz Thoma$^{2}$\\Maximilian Forstenh\"ausler$^{2,3}$, Nael Fasfous$^{2,\dagger}$, Manoj Rohit Vemparala$^{2}$, Pierpaolo Mori$^{2}$\\Christos Anagnostopoulos$^{3}$, Daniel Mueller-Gritschneder$^{4}$, Ulf Schlichtmann$^{1}$\\[3pt]
{\small $^{1}$Technical University of Munich, Germany \enspace $^{2}$BMW Group, Germany \enspace $^{3}$University of Glasgow, UK \enspace $^{4}$TU Wien, Austria}\\[1pt]
{\tt\footnotesize \{firstname.lastname\}@\{$^{1}$tum.de, $^{2}$bmw.de, $^{3}$glasgow.ac.uk, $^{4}$tuwien.ac.at\}}
}

\begin{document}
\maketitle
{\renewcommand{\thefootnote}{*}\footnotetext{Equal contribution. Co-first authors are ordered in alphabetical order.}}
{\renewcommand{\thefootnote}{$\dagger$}\footnotetext{Employed at BMW at time of submission.}}

\begin{abstract}
The integration of vision-language models (VLMs) into end-to-end (E2E) autonomous driving has demonstrated strong potential by providing scene description and reasoning capabilities. However, VLMs are computationally expensive, and switching language reasoning on and off at runtime without hurting closed-loop performance has not been demonstrated, which hinders practical adoption. To address these issues, we propose DRIFT, an E2E autonomous driving framework that integrates a lightweight language model (LM) for visual question answering and action commentary. By efficiently projecting intermediate driving representations into the language embedding space, we demonstrate that a compact LM ($<$150M parameters) can achieve strong scene description and action commentary. Furthermore, we introduce a runtime-flexible planning module that is trained on surrogate reasoning features distilled from the LM and can be conditioned on either the LM or the surrogate reasoning features at inference. This design enables bypassing the LM entirely when required by deployment constraints, without sacrificing driving performance. 
With only 255M parameters, DRIFT reaches up to a Driving Score (DS) of 97.03 and a Success Rate of 89.39\% (mean over three runs) on Bench2Drive by interleaving its LM and surrogate reasoning at a chosen reasoning rate. This is on par with the strongest non-language models while being far more compact than existing VLM-based models. DRIFT further achieves competitive results on the real-world NAVSIM and AlpaSim benchmarks at significantly lower computational cost than existing VLM-based approaches. Project page: \mbox{\href{https://drift-e2e.github.io/}{\nolinkurl{drift-e2e.github.io}}}
\end{abstract}

\section{Introduction}

End-to-end (E2E) driving models \cite{hu2023planning, jiang2023vad} achieve promising driving performance, but they lack interpretability and high-level semantic reasoning. To address these limitations, recent works have integrated vision-language models (VLMs) into the driving pipeline, enabling visual question answering (VQA) \cite{sima2024drivelm, tian2024drivevlm} and language-guided reasoning \cite{sima2024drivelm, fu2025orion}.
Despite their success, existing VLM-based E2E methods typically improve language performance by scaling up the language models (LMs), incurring substantial computational overhead. Approaches such as DriveVLM \cite{tian2024drivevlm}, ORION \cite{fu2025orion}, and AdaThinkDrive~\cite{luo2025adathinkdrive} employ models exceeding 7B parameters, making real-time deployment on edge hardware increasingly challenging. This motivates us to study how strong language reasoning can be achieved with a significantly smaller LM.

A second limitation is that these frameworks couple language reasoning tightly to the driving pipeline, requiring LM inference throughout deployment. To reduce this cost, some models run separate fast (standard E2E) and slow (VLM-based) branches and switch between them at runtime~\cite{qian2025fasionad++, hamdan2025eta, zhang2025adadrive}, while others switch a single VLA between thinking and non-thinking modes~\cite{luo2025adathinkdrive, zhou2026autovla}, but most of them still rely on a large model in complex scenes. Others distill the LM into the driving model, such as DiMA~\cite{hegde2025distilling}, VLM-AD~\cite{xu2025vlmad}, and VERDI~\cite{feng2025verdi}, but language only shapes the planner during training, and the planner cannot take reasoning as input at inference. DiMA's Dual variant is the only one of them that is able to run the LM at inference, but it is evaluated only in open-loop settings, leaving the switching of the LM on and off in difficult closed-loop driving unexplored.

We argue that efficiently integrating a language model into E2E driving requires two properties: \emph{compactness}, so that the LM adds little computational cost, and \emph{detachability}, so that the LM can be removed at runtime without degrading driving score. We propose DRIFT, an E2E driving framework that realizes this paradigm.

Our main contributions are summarized as follows:
\begin{itemize}
\item Efficient driving VQA and action commentary with a compact LM: by aligning dense driving representations with the language embedding space, a 135M-parameter LM achieves competitive VQA and action commentary at an edge-compatible latency on a Jetson AGX Orin. Additionally, as an interesting finding, we show that language training also enriches the shared backbone features.
\item Detachable LM reasoning: we distill the LM's reasoning into a lightweight surrogate module, so that a single planner can run on either reasoning source and the LM can be enabled or detached without degrading driving score.
\item Extensive closed-loop experiments on Bench2Drive show, first, that language training improves driving over the language-free base model (+0.5 DS), and second, that combining the LM and surrogate reasoning at a chosen reasoning rate reaches 97.03 DS (+1.8 over TFv6/LEAD~\cite{nguyen2026lead}). On the real-world NAVSIM benchmark, DRIFT is competitive with much larger VLM-based planners, and on AlpaSim it is comparable to the LTFv6 and TransFuser-K1 baselines. Moreover, similar to Bench2Drive, enabling the LM on these benchmarks does not degrade planning performance, which shows that the detachable design carries over from simulation to real-world data.
\end{itemize}
\section{Related work}

\subsection{VLMs in end-to-end autonomous driving}
Pioneering E2E autonomous driving models \cite{jiang2023vad, jaeger2023hidden, zimmerlin2024hidden} show promising results in directly mapping raw sensors to actions. Recently, LEAD~\cite{nguyen2026lead} showed that reducing the asymmetry between student and expert boosts performance, yielding the strong TFv6 and the diffusion-based BridgeDrive~\cite{liu2026bridgedrive} models. However, despite their success, these models offer limited interpretability and lack explicit language reasoning. Hint-AD~\cite{ding2024hint} improves interpretability by decoding language from the intermediate features of a frozen E2E model, but its language only explains the model's decisions post hoc and does not influence driving. This motivates the integration of VLMs into E2E driving.

VLM-based autonomous driving frameworks \cite{lu2025real, marcu2024lingoqa, zhou2026opendrivevla, zeng2026futuresightdrive, xie2026latentvla, li2026sgdrive, wang2025unified, zhou2026qwen} show the potential of integrating LMs into E2E driving through VQA and language-guided driving. However, they were primarily evaluated in open-loop settings, which may not reflect real-world driving \cite{li2024ego, jia2024bench2drive}. This has motivated closed-loop VLM-based driving \cite{yang2026drivemoe, fu2025orion, kong2025vlr, zhao2026vladriver, fu2025minddrive}, where SimLingo \cite{renz2025simlingo} shows that a 1B model suffices when reasoning and driving actions are generated jointly, while LinkVLA \cite{wang2026unifying} and FIVE-VLA \cite{oksuz2026five} further improve efficiency. Yet these frameworks are computationally demanding because of their large LMs, the tight coupling between language reasoning and planning, or both, which limits deployment.

\subsection{Efficient reasoning for autonomous driving}

\paragraph{Fast-slow methods.} To reduce the inference cost of VLMs, fast-slow methods invoke them only when needed. Dual-branch designs~\cite{tian2024drivevlm, qian2025fasionad++, hamdan2025eta, zhang2025adadrive, li2026think} pair a fast VLM-free path with a slow VLM path, while adaptive-thinking VLAs~\cite{luo2025adathinkdrive, zhou2026autovla, ling2026blue} switch a single model between thinking and direct action. However, most of them still rely on a large VLM or VLA, whose use remains mandatory in complex scenes, which limits deployment under resource constraints. DRIFT also invokes its LM at a chosen reasoning rate, but with a compact LM, a single planner, and the option to bypass the LM without a drop in driving performance.

\paragraph{VLM knowledge distillation.} A complementary direction transfers VLM knowledge into the driving model during training so that the VLM can be dropped at deployment. VLM-AD \cite{xu2025vlmad} and VERDI \cite{feng2025verdi} use an offline VLM to do reasoning on training samples and train the planner's features to match text embeddings of it, whereas DiMA \cite{hegde2025distilling} jointly trains a large multi-modal LM (Vicuna 7B) with the planner and aligns their hidden features. In all three, language only shapes the planner during training and the planner cannot take reasoning as input at inference. DiMA's Dual variant can also run the LM at inference by fusing LM and planner features, but these methods are evaluated mostly open-loop or in simplified closed-loop settings, so, to our knowledge, switching the LM on and off while driving was not evaluated. In DRIFT, a single planner works interchangeably and flexibly with both surrogate and LM reasoning sources in challenging closed-loop driving.
\section{Proposed method: DRIFT}
\begin{figure*}[t]
    \centering
    \includegraphics[width=0.58\textwidth]{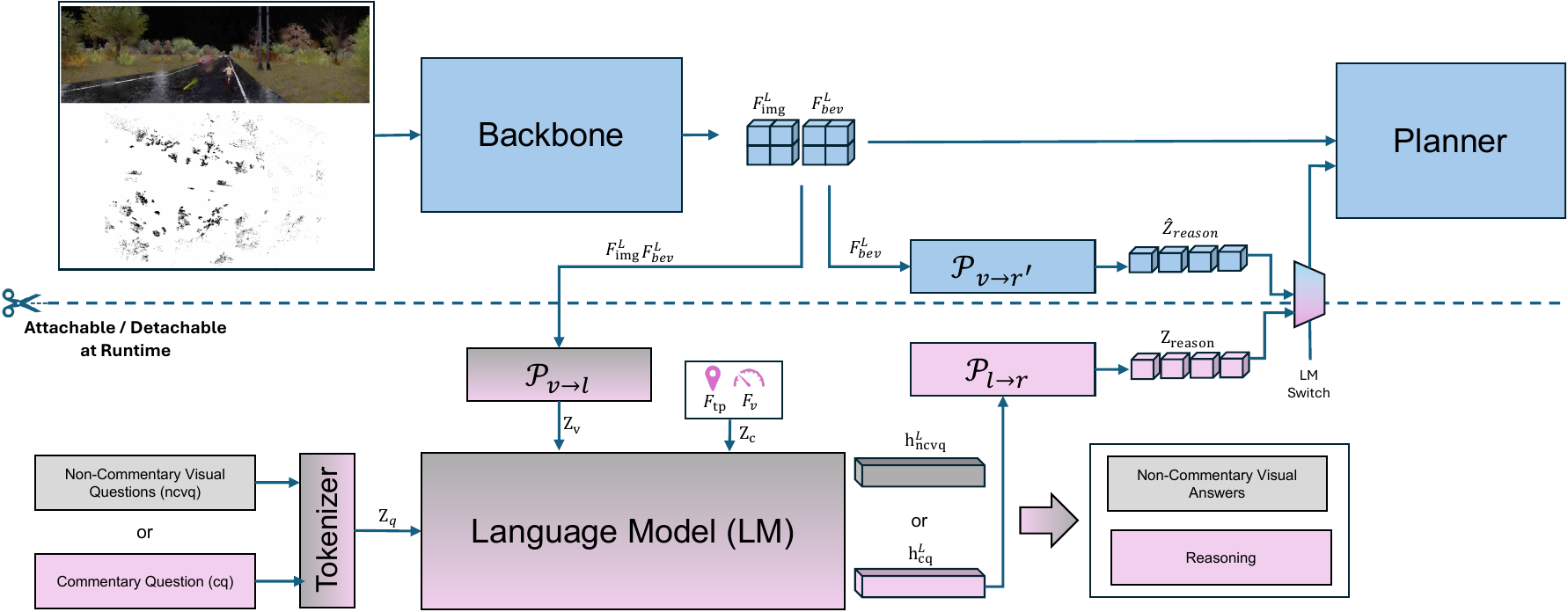}
    \hfill
    \includegraphics[width=0.41\textwidth]{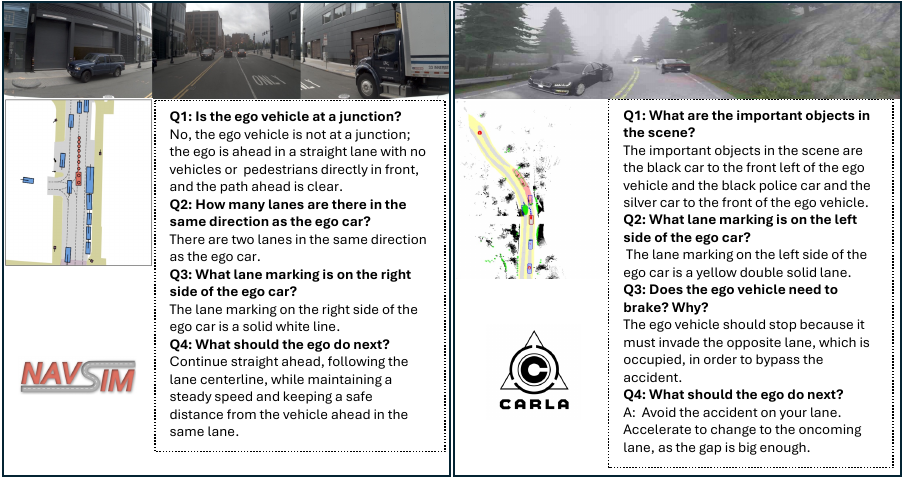}
    \caption{Overview of our proposed method architecture (left) and qualitative driving results (right).}
    \label{fig:full_overview}
\end{figure*}

In this section, we present DRIFT, an E2E autonomous driving framework equipped with a lightweight, plug-and-play language reasoning module. An overview of the proposed architecture is shown in \cref{fig:full_overview}. Given raw sensor observations, a TransFuser-based backbone extracts a dense scene representation, which serves as the input to the LM for VQA (\cref{sec:proposed_method_vqa}). In addition, the planner is conditioned on surrogate reasoning tokens that are aligned to the reasoning tokens of the LM, which can optionally replace them at inference (\cref{sec:proposed_method_dereason}), to facilitate aligned reasoning and driving decisions. Finally, \cref{sec:proposed_method_tr_proc} describes the training procedure that enables seamless integration of the detachable language module without degrading driving performance.

\subsection{Efficient visual question answering}
\label{sec:proposed_method_vqa}
A core component of DRIFT is an efficient VQA module that enables a compact LM to interpret driving scenes without introducing substantial computational overhead. To design an effective interface between the perception backbone and the language module, we investigate two fundamental questions: (1) which scene representations are most beneficial for language understanding, and (2) how these representations should be projected into the language embedding space.
Given raw inputs $X$, the perception backbone first extracts dense scene representations:
\begin{equation}
\left(F_{img}^{1:L},\,F_{bev}^{1:L}\right)=\Phi_{\theta}(X),
\end{equation}

where $\Phi_{\theta}(\cdot)$ denotes the TransFuser-based perception backbone and $L$ is the number of backbone layers. Similar to TOKEN~\cite{tian2024tokenize}, we decode language from the driving backbone's representations rather than from a separate vision encoder. However, instead of object-level tokens,  we use the dense BEV and image features extracted from the vision backbone's final layer (denoted as $F_{img}^{L}$ and $F_{bev}^{L}$ respectively), which our studies show to provide the most effective visual representation for language tasks (shown in \cref{tab:feature_selection}). To transform these features into a language-compatible space, we investigate different projectors, including attention-based feature aggregation. Our experiments show that a lightweight per-token MLP is sufficient and achieves better performance than these more complex alternatives (shown in \cref{tab:feature_selection}). Therefore, we employ a lightweight per-token projector $\mathcal{P}_{v\rightarrow l}$ (a two-layer MLP) that maps the visual features into the embedding space of the LM. Both feature maps have the same channel width of 1512, with $F_{img}^{L}\in\mathbb{R}^{432\times 1512}$ (a $12{\times}36$ grid) and $F_{bev}^{L}\in\mathbb{R}^{120\times 1512}$ (a $10{\times}12$ grid), so they are concatenated along the token axis into 552 tokens without a channel-matching projection, producing visual tokens $Z_v\in\mathbb{R}^{552\times d}$ with the LM embedding size $d$:
\begin{equation}
Z_v = \mathcal{P}_{v\rightarrow l}\!\left(\left[F_{img}^{L};\ F_{bev}^{L}\right]\right).
\end{equation}
In addition, we incorporate driving context to provide the LM with essential motion and navigation cues. The scalar ego-speed $v_e$ is projected through an MLP to obtain speed features $F_{v}$. Similarly, the target point is processed by an MLP to produce target-point features $F_{tp}$. These are concatenated to form the context tokens $Z_c$:
\begin{equation}
Z_c = \left[F_{v};\ F_{tp}\right].
\end{equation}
For a given question \(Q\in \mathcal{Q}_{ncvq}\cup \mathcal{Q}_{cq}\) (comprising non-commentary visual questions \(\mathcal{Q}_{ncvq}\) and commentary questions \(\mathcal{Q}_{cq}\)), we tokenize and embed the input query into continuous question token embeddings \(Z_{q}=\text{Embed}(\text{Tokenizer}(Q))\). Finally, the visual, context, and question tokens are concatenated and provided to the compact LM for VQA:
\begin{equation}
Z_{input} = [Z_v;\ Z_c;\ Z_q].
\end{equation}

\subsection{Detachable reasoning}
\label{sec:proposed_method_dereason}

Effective reasoning for autonomous driving requires alignment between language reasoning and driving decisions, as also highlighted in \cite{renz2025simlingo}. At the same time, we want the language module to be optional at runtime. We address both with a set of \emph{surrogate} reasoning tokens that condition the planner together with dense scene features. They are predicted from BEV features alone, which makes the planner independent of the LM. The LM serves as the teacher during training, and as an optional replacement for the surrogate module at inference, allowing the planner to condition on language reasoning.

Concretely, we use the generated commentary response as the reasoning signal, since it describes the current action together with its justification. During the final training stage and inference, the reasoning tokens are computed from the LM's own greedy commentary, not from ground-truth text. We extract the last-layer hidden states at the generated answer positions, denoted as $h_{cq}^L$, and pool them into $K$ fixed-size reasoning tokens $Z_{reason}$ through an attention pooling $\mathcal{P}_{l\rightarrow r}$, in which $K{=}16$ queries attend over the variable-length hidden states with padded positions masked. $\mathcal{P}_{l\rightarrow r}$ is randomly initialized and kept fixed, i.e., it receives no gradient, so $Z_{reason}$ is a fixed random-feature view of the LM hidden states, which are themselves trained by $\mathcal{L}_{lm}$. In parallel, a lightweight surrogate projector $\mathcal{P}_{v\rightarrow \tilde{r}}$ (a $1{\times}1$ convolution followed by an attention pooling of the same form over the BEV grid) predicts $K$ surrogate reasoning tokens of the same shape from the BEV features $F_{bev}^{L}$ alone, $\tilde{Z}_{reason} = \mathcal{P}_{v\rightarrow \tilde{r}}(F_{bev}^{L})$. During training, the planner is conditioned on the last-layer BEV features together with the surrogate tokens, and the planning loss is applied only to this path. The LM tokens $Z_{reason}$ do not supervise the planner. Instead, they serve as detached targets (the alignment losses do not backpropagate into the LM or into $\mathcal{P}_{l\rightarrow r}$) that the surrogate tokens learn to mimic through a reasoning distillation (\cref{sec:proposed_method_tr_proc}), so that both token types live in the same representation space. We also tried feeding the LM tokens to the planner during training, i.e., additionally optimizing the planning loss on the LM path, but this degraded driving performance (see the supplementary), so the planner is trained on the surrogate tokens only.

At inference, the same planner can be conditioned on either token type (\cref{eq:plan_ctx}). Because the surrogate reasoning tokens are trained to be aligned to the LM reasoning tokens, the planner can also be conditioned on the LM tokens without retraining (\cref{tab:recipe_short}). Consequently, the language module can be detached completely based on user preference, or invoked at a chosen reasoning rate (\cref{sec:insights}) with the surrogate tokens used in between, without impacting driving performance. In the planning head, we construct a unified context $Z_{\text{plan}}$ by concatenating the flattened BEV tokens (equipped with 2D cosine positional embeddings), ego status tokens $Z_{\text{status}}$, and the reasoning tokens $Z_{\text{r}}$:
{\small
\begin{equation}
    Z_{\text{plan}} = \left[ F_{\text{bev}}^L ; Z_{\text{status}} ; Z_{\text{r}} \right],\,
    Z_{\text{r}} = \begin{cases} \tilde{Z}_{\text{reason}} & \text{LM detached},\\ Z_{\text{reason}} & \text{LM invoked}. \end{cases}
\label{eq:plan_ctx}
\end{equation}\vspace{0pt}}%
\noindent Hence, $Z_{\text{plan}}$ is the only planner context. Architecture details are in the supplementary. During training, the planner always receives the surrogate tokens, i.e., $Z_{\text{r}} = \tilde{Z}_{\text{reason}}$. The LM is trained by its own language loss, and its tokens only serve as alignment targets. At inference, $Z_{\text{r}}$ is determined by whether the LM is detached or invoked. The status tokens $Z_{\text{status}}$ encapsulate ego-vehicle states (e.g., velocity, navigation commands, and target points) projected via linear encoders and differentiated through per-slot status positional embeddings $E_{\text{status}}$. Finally, a Transformer decoder queries $Z_{\text{plan}}$ using learned query vectors $Q_{\text{plan}}$, $h_{\text{dec}} = \text{Decoder}(Q_{\text{plan}}, Z_{\text{plan}})$. Linear heads consume $h_{\text{dec}}$ to predict spatial path, temporal waypoints, and target speed distributions. In closed-loop inference, we use the predicted path for steering and the speed distribution for throttle and brake selection.

\subsection{Training}
\label{sec:proposed_method_tr_proc}
\subsubsection{Training objective}

Our overall training objective is defined in \cref{eq:tr_loss}. To learn rich scene representations, we first optimize a set of perception tasks, including depth estimation, semantic segmentation in both RGB and BEV views, and object detection. Their corresponding losses are jointly denoted as $\mathcal{L}_{\text{perc}}$. The language model is trained to autoregressively predict VQA answer tokens using the standard cross-entropy loss, denoted by $\mathcal{L}_{\text{lm}}$.
For planning, we supervise the predicted route using an L1 loss over route points ($\mathcal{L}_{\text{cp}}$), an L1 loss over future ego waypoints ($\mathcal{L}_{\text{wp}}$), and a cross-entropy loss over target speed bins ($\mathcal{L}_{\text{ts}}$), which together define the planning loss $\mathcal{L}_{\text{plan}}=\mathcal{L}_{\text{cp}}+\mathcal{L}_{\text{wp}}+\mathcal{L}_{\text{ts}}$. It is applied to the surrogate planning branch only, yielding $\mathcal{L}_{\text{plan}}^{\text{surr}}$. Additionally, two alignment terms pull the surrogate tokens toward the detached LM tokens: a pointwise MSE $\mathcal{L}_{\text{sim}}=\mathrm{MSE}\bigl(\tilde{Z}_{\text{reason}}, \mathrm{sg}(Z_{\text{reason}})\bigr),$ with stop-gradient $\mathrm{sg}(\cdot)$, and a relational KL loss $\mathcal{L}_{\text{con}}$ that matches the cosine-similarity distribution of the mean-pooled tokens across samples (temperature $\tau=0.1$; formula in the supplementary). The alignment terms are gradually introduced using $\alpha(t)=\min(1, t/0.3T)$, where $t$ and $T$ are the current and total numbers of training steps of the final stage (Stage~3b, \cref{sec:proposed_method_schedule}), which linearly increases from $0$ to $1$ during the first $30\%$ of that stage. This warm-up lets the surrogate projector, which already imitates the ground-truth reasoning tokens from Stage~2, adapt gradually to the LM's own (on-policy) tokens. The overall objective is
{\small
\begin{equation}
    \mathcal{L} = \mathcal{L}_{\text{perc}} + \mathcal{L}_{\text{lm}} + \mathcal{L}_{\text{plan}}^{\text{surr}} + \alpha(t)\,(\lambda_{\text{sim}}\mathcal{L}_{\text{sim}} + \lambda_{\text{con}}\mathcal{L}_{\text{con}}),
\label{eq:tr_loss}
\end{equation}}
where $\mathcal{L}_{\text{perc}}$ is a weighted sum of its terms; all loss weights are in the supplementary.

\subsubsection{Training schedule}
\label{sec:proposed_method_schedule}
Motivated by the number of modules and their dependencies, we split training into three stages. Stage~1 trains the backbone and perception heads to produce the visual features all downstream modules rely on. Stage~2 trains the language modules (the LM with LoRA, the projector, and the surrogate projector) to answer VQAs and reason in language, as well as the auxiliary perception tasks of Stage~1, with the alignment losses active on the ground-truth commentary (teacher-forced), while the backbone receives the LM gradient scaled down by 0.1. Stage~3 is the final end-to-end training. Stage~3a trains the backbone, planner, and surrogate path on all frames without the LM. Stage~3b additionally trains the LM jointly: the alignment losses are re-introduced with reduced weights and a warm-up, and are computed on the LM's own greedy commentary, which matches inference and avoids targets that depend on the expert's future, at the cost of one extra LM pass per step (the language loss uses the teacher-forced pass). Ground-truth text is used earlier because it is cheap and informative while the LM is still weak.

\section{Experiments}
\label{sec:new_experiments}

\subsection{Experimental setup}
We evaluate DRIFT in closed loop on the challenging Bench2Drive benchmark~\cite{jia2024bench2drive} in the CARLA simulator~\cite{dosovitskiy2017carla}, reporting Driving Score (DS) and Success Rate (SR). We focus on closed-loop evaluation, since the planner's actions affect subsequent states, which reflects driving behavior better than open-loop evaluation~\cite{renz2025simlingo,zimmerlin2024hidden}. Training uses the LEAD dataset~\cite{nguyen2026lead}, for which we generate VQA annotations following DriveLM~\cite{sima2024drivelm} and commentary annotations following SimLingo~\cite{renz2025simlingo}; language quality is measured with SPICE, CIDEr, BLEU-4, ROUGE-L and METEOR. To evaluate DRIFT on real-world data, we train it on the navtrain split of NAVSIM~\cite{dauner2024navsim, cao2025pseudo} and report the PDM Score (PDMS) on navtest and the extended PDM Score (EPDMS) on navhard. We additionally evaluate it in closed loop on the nuPlan track of AlpaSim~\cite{alpasim_2025}, following the 123D protocol~\cite{dauner2026123d}. Language annotations for NAVSIM are generated with Qwen3-VL~\cite{bai2025qwen3}. Training consists of three stages, namely perception pre-training following TFv6~\cite{nguyen2026lead}, LoRA~\cite{hu2022lora} finetuning of the LM with the alignment losses, and end-to-end training, and runs on 8 NVIDIA H100 GPUs. Further details on the data, metrics and training are in the supplementary.

\subsection{Main results}
We compare DRIFT with prior language-based models in \cref{tab:lm_reasoning_rate_new}. Benefiting from the finding of LEAD~\cite{nguyen2026lead} that reducing the learner--expert asymmetry improves imitation learning, DRIFT is trained on the data collected by LEAD. Some works use additionally collected data, which results in inconsistent training data across methods, from the 234K frames of the Bench2Drive training set to 3.1M frames for SimLingo, and some do not report it (see the column in the table). Hence, the numbers are not strictly comparable, and we read them as a comparison of complete driving recipes, i.e., data, backbone, and architecture together. In this comparison, DRIFT w/ LM improves over the strongest language-based model, LinkVLA~\cite{wang2026unifying}, by 6.0 DS and 14.8 SR with about $4\times$ fewer parameters (255M vs.\ 1B). Moreover, SimLingo~\cite{renz2025simlingo} reports that its language-free version, SimLingo-BASE, drives as well as the full model, i.e., language preserves but does not improve its driving performance, whereas in DRIFT the LM can improve it within a single model.

Next, we compare DRIFT with and without language invocation to the non-language models TFv6~\cite{nguyen2026lead} and BridgeDrive~\cite{liu2026bridgedrive}, which are trained on the same LEAD data. Without the LM, DRIFT is on par with TFv6 (DS 95.97 vs.\ 95.2). Invoking the LM then adds 1.1 DS, giving the best result on Bench2Drive: DRIFT w/ LM reaches 97.03 DS and 89.39 SR, ahead of TFv6 by 1.8 DS and 2.6 SR. BridgeDrive is a diffusion-based model with a more complex planning head. Against it, the margin is smaller (+0.7 DS) and mostly within one standard deviation of the evaluation runs, making DRIFT slightly ahead in DS.

\begin{table}[t]
    \centering
    \caption{\textbf{Closed-loop results on Bench2Drive.} We compare DRIFT with language-based and non-language models in Driving Score (DS) and Success Rate (SR). Frames: size of the training set (``-'': not reported). DRIFT results are mean$_{\pm\text{std}}$ over 3 closed-loop runs (sample standard deviation); DRIFT w/ LM uses a reasoning rate of 2.}
    \label{tab:lm_reasoning_rate_new}
    \footnotesize
    \setlength{\tabcolsep}{3.5pt}
    \resizebox{\linewidth}{!}{%
    \begin{tabular}{l c c c c}
        \toprule
        Method & Params & Frames & DS $\uparrow$ & SR (\%) $\uparrow$ \\
        \midrule
        \multicolumn{5}{l}{\textit{Language-based models}} \\
        DriveMoE \cite{yang2026drivemoe} & 3B & 234K & 74.22 & 48.64 \\
        ETA \cite{hamdan2025eta} & 7B & 234K & $74.33_{\pm 0.96}$ & $48.33_{\pm 3.56}$ \\
        VLR-Driver \cite{kong2025vlr} & 8.1B & 234K & 75.01 & 50.00 \\
ORION \cite{fu2025orion} & 7B & 234K & 77.74 & 54.62 \\
        SpaceDrive+ \cite{li2026spacedrive} & 7.4B & - & 78.02 & 55.11 \\
        AutoVLA \cite{zhou2026autovla} & 3B & 600K & 78.84 & 57.73 \\
        MindDrive-L \cite{fu2025minddrive} & 3B & 234K & 80.59 & 58.26 \\
        SimLingo \cite{renz2025simlingo} & 1B & 3.1M & $85.07_{\pm 0.95}$ & $67.27_{\pm 2.11}$ \\
        VLADriver-RAG \cite{zhao2026vladriver} & - & 234K & 89.12 & 70.42 \\
        LinkVLA \cite{wang2026unifying} & 1B & - & 91.01 & 74.55 \\
        \midrule
        \multicolumn{5}{l}{\textit{Non-language models}} \\
        SimLingo-BASE \cite{renz2025simlingo} & 355M & 3.1M & 85.94 & 66.82 \\
        TFv6 \cite{nguyen2026lead} & 126M & 1M & $95.2_{\pm 0.3}$ & $86.8_{\pm 0.7}$ \\
         BridgeDrive \cite{liu2026bridgedrive} & - & 1M & 96.34$_{\pm 0.55}$ & 89.25$_{\pm 0.50}$ \\
        \midrule
        \textbf{DRIFT w/o LM }& 120M & 1M & $95.97_{\pm 0.65}$ & $86.52_{\pm 1.60}$ \\
        \textbf{DRIFT w/ LM} & 255M & 1M & \textbf{97.03}$_{\pm 0.09}$ & \textbf{89.39}$_{\pm 0.95}$\\
        \bottomrule
    \end{tabular}%
    }
\end{table}

\begin{table}[t]
  \centering
  \caption{Language evaluation results. B-4 = BLEU-4, S = SPICE, C = CIDEr, R-L = ROUGE-L, M = METEOR.}
  \label{tab:drivelm_vqa}
  \resizebox{\columnwidth}{!}{
  \begin{tabular}{@{}llccccc@{}}
    \toprule
    \textbf{Dataset} & \textbf{Model} & \textbf{B-4}$\uparrow$ & \textbf{S}$\uparrow$ & \textbf{C}$\uparrow$ & \textbf{R-L}$\uparrow$ & \textbf{M}$\uparrow$ \\
    \midrule
    \multirow{3}{*}{DriveLM-CARLA}
      & DriveLM-CARLA (LLaMA-7B) & - & 72.71 & 1.15 & 71.20 & - \\
      & SimLingo (Qwen2-0.5B)    & 72.07 & \textbf{87.24} & \textbf{7.08} &  84.77 & 50.57 \\
      & Ours (SmolLM2-135M)      & \textbf{72.63} & 85.56 & 6.96 & 84.59 & 50.08 \\
    \midrule
    Commentary & Ours (SmolLM2-135M) & 83.90 & 82.20 & 6.93 & 89.10 & 57.01 \\
    \bottomrule
  \end{tabular}}
\end{table}

Besides competitive driving performance, DRIFT also produces high-quality language (\cref{tab:drivelm_vqa}). On DriveLM-CARLA, DRIFT with a 135M-parameter LM achieves performance comparable to SimLingo (0.5B). Both are evaluated zero-shot there after being trained on separate datasets. We attribute this efficiency to our feature-selection and projector designs. DRIFT decodes language directly from the driving backbone's intermediate representations. Because these features already encode rich, planning-oriented information on the driving scenario, a small model is sufficient for language decoding in the form of structured driving VQA and commentary.

Crucially, the language head and the planner consume the same backbone features, so the LM's outputs are grounded in the representation used for driving, which makes the generated language suitable as a reasoning signal for the planner. The commentary response, which states the intended action together with its justification, is reliable enough to serve both as the target that the surrogate reasoning tokens distill from during training, and as the planner's conditioning when the LM is invoked at inference. The commentary answers are more demanding than DriveLM answers, since they must state the intended action together with its justification instead of only describing the scene. As shown in \cref{tab:drivelm_vqa}, they still reach high quality (SPICE 82.20), confirming their suitability as the reasoning signal.

\subsection{Insights}
\label{sec:insights}
\begin{figure}[t]
    \centering
    \includegraphics[width=\columnwidth]{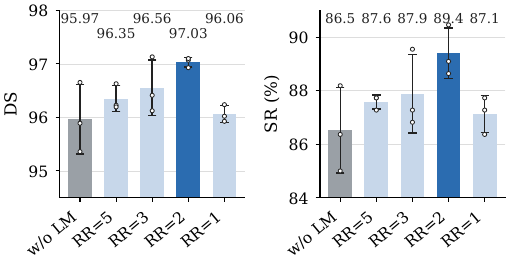}
    \caption{\textbf{Effect of the LM reasoning rate (RR) on Bench2Drive.} Bars show the mean and error bars the sample standard deviation over 3 closed-loop runs; dots are individual runs. The best setting (RR=2, selected on this benchmark) is highlighted.}
    \label{fig:rr_rate}
\end{figure}

\paragraph{How often should the LM reason?}
Because the LM in DRIFT is detachable, we can study how often reasoning needs to be invoked. We vary the reasoning rate (RR), i.e., the LM is invoked every RR frames and the surrogate reasoning is used in between (\cref{fig:rr_rate}), and make two observations. First, invoking the LM periodically slightly improves driving over the model without LM (95.97 DS), with the best setting, RR=2, reaching 97.03 DS (+1.06 over no LM, 95\% route-level bootstrap CI [0.2, 2.0]). Since we selected RR on this benchmark, we regard the gain as indicative. The gains of the other rates are within run-to-run variation. Second, invoking the LM at every frame (RR=1) is on par with no reasoning (96.06 vs.\ 95.97 DS) and below the intermediate rates. We use a fixed reasoning rate, which gives predictable latency at runtime. Because surrogate and LM tokens are interchangeable, the switch between them could also be controlled by an adaptive trigger; since a trigger brings its own design choices and failure modes that could confound the impact of DRIFT, we leave studying adaptive strategies to future work.

\begin{figure}[t]
  \centering
  \setlength{\tabcolsep}{1.5pt}
  \begin{tabular}{@{}c c c@{}}
    \footnotesize Input & \footnotesize LM path & \footnotesize Surrogate path \\[2pt]
    \includegraphics[width=0.32\linewidth]{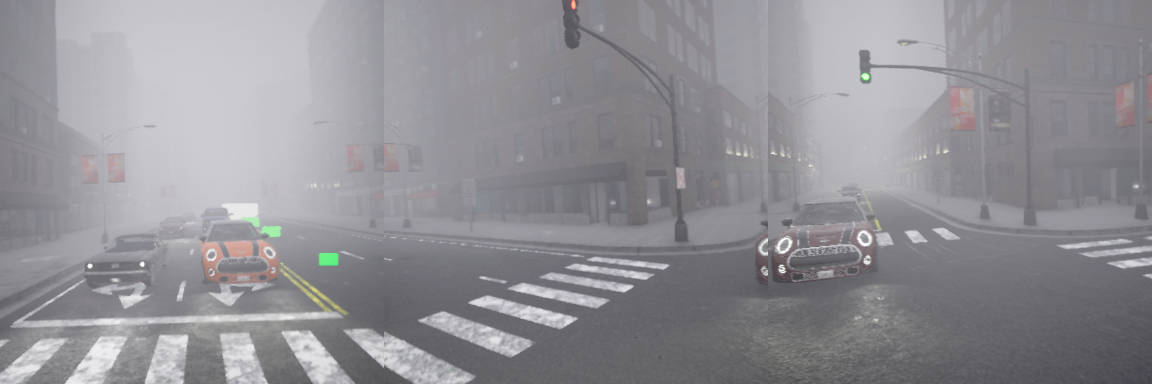} &
    \includegraphics[width=0.32\linewidth]{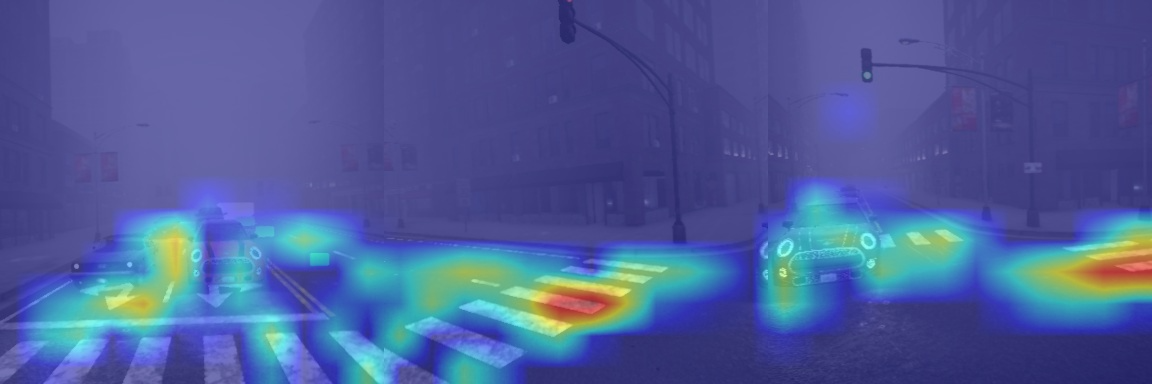} &
    \includegraphics[width=0.32\linewidth]{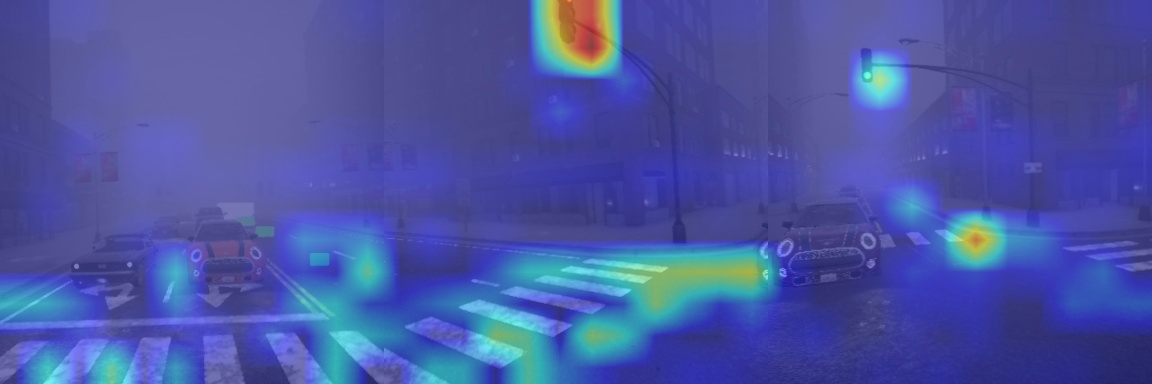} \\
  \end{tabular}
      \caption{Grad-CAM maps for the LM and surrogate paths in a junction left turn scenario. While the LM maintains broader attention across all critical scene objects, the surrogate path focuses more narrowly on driving-relevant segments.}
  \label{fig:attribution}
\end{figure}

To understand why the LM and surrogate paths differ, we visualize the Grad-CAM heatmaps of the reasoning tokens produced by the LM and surrogate paths for an unprotected junction left turn scenario (\cref{fig:attribution}). The surrogate branch mainly focuses on the ego lane and a few localized cues, such as the traffic light. In contrast, the language reasoning branch shows broader contextual awareness, attending not only to the ego lane but also to semantically relevant cues such as the oncoming vehicle and the pedestrian crossings.

These observations suggest that the two reasoning paths provide complementary information. The LM contributes richer scene-level reasoning for complex decisions, while the surrogate emphasizes localized cues that suffice for most planning updates. Consistently, the gain of RR=2 concentrates in negotiation scenarios, e.g., unprotected left turns (DS 84.5 without LM vs.\ 96.9 at RR=2 on 15 routes). Refreshing language reasoning at a moderate rate thus lets DRIFT benefit from semantic reasoning without overriding the stable surrogate reasoning at every step. More analysis is in the supplementary.

\paragraph{Is DRIFT's reasoning aligned with its actions?}
We measure how well the LM's textual reasoning agrees with the action taken by the planner in DRIFT, with the LM invoked at every frame (RR=1). As a baseline we use a decoupled variant of DRIFT, in which the LM is trained on the same commentary data but the planner is never conditioned on its reasoning. Since DRIFT predicts path and speed separately, we evaluate the lateral and longitudinal axes separately, both in closed loop on Bench2Drive and in open loop on NAVSIM (\cref{tab:faithfulness}). The evaluation protocol and the decoupled models are described in the supplementary.

Both variants are far above the constant-answer reference. The decoupled variant is already well aligned, which is expected since the commentary and the planner's ground truth come from the same expert. Furthermore, linking the planner to the LM's reasoning increases longitudinal agreement by 6.0 points on Bench2Drive (93.2 vs.\ 87.2) and 3.4 points on NAVSIM (79.4 vs.\ 76.0), where planning is highly multi-modal~\cite{zimmerlin2024hidden}. Laterally, both variants agree equally (85.2 vs.\ 85.0 and 89.4 vs.\ 88.9), as lateral decisions are largely determined by the road geometry~\cite{zimmerlin2024hidden}. While both benchmarks show similar trends, the numbers should not be compared directly, as their evaluation protocols differ. Lateral agreement is a coarse measure, since several descriptions can be correct for the same situation, so 85\% reflects a limit of the evaluation rather than low alignment.

\begin{table}[t]
    \centering
    \caption{\textbf{Reasoning--action alignment.} Agreement (\%) between DRIFT's generated reasoning and its plan on Bench2Drive (closed loop) and NAVSIM (open loop). Decoupled: same language model trained on the same commentary, but the planner never reads its reasoning. Constant answer: best score from always giving the same answer. Protocol details in the supplementary.}
    \label{tab:faithfulness}
    \footnotesize
    \setlength{\tabcolsep}{3pt}
    \begin{tabular}{l c c c}
        \toprule
         & DRIFT & Decoupled & Constant answer \\
        \midrule
        \multicolumn{4}{l}{\textit{Bench2Drive (closed loop)}} \\
        Longitudinal & \textbf{93.2} & 87.2 & 65.6 \\
        Lateral & \textbf{85.2} & 85.0 & 73.9 \\
        \midrule
        \multicolumn{4}{l}{\textit{NAVSIM (open loop)}} \\
        Longitudinal & \textbf{79.4} & 76.0 & 33.3 \\
        Lateral & \textbf{89.4} & 88.9 & 33.3 \\
        \bottomrule
    \end{tabular}
\end{table}

\subsection{Embedded inference efficiency}
\label{sec:embedded_efficiency}

We evaluate embedded inference on an NVIDIA Jetson AGX Orin in MAXN power mode at batch size one (\cref{tab:embedded_efficiency}). The backbone and planner use static-shape TensorRT engines, and the language model runs in TensorRT Edge-LLM. LoRA weights are merged in FP32 before FP16 export. Numerically sensitive backbone attention, normalization, and reductions remain in FP32.

We adapt Edge-LLM to accept projected multimodal embeddings and return final-layer hidden states for generated answer tokens. Attention pooling reduces these states to fixed-size reasoning tokens for the planner, keeping intermediate features on the GPU. KV caching avoids repeated prompt computation, while reusing answer states avoids full-sequence replay when token alignment permits. Language reasoning runs at every planning step when enabled.
SimLingo uses the same TensorRT and Edge-LLM approach, with BF16 driving heads.
Latency is averaged over closed-loop inference calls with varying scenes and generated answer lengths. Peak NvMap GPU allocation and logical FLOPs (FMA${}=2$) use separate fixed-input profiles. Sampled peak GPU allocation is 0.48 GiB with the LM unloaded and 1.69 GiB with it resident for runtime switching.

\begin{table}[t]
\centering
\caption{\textbf{Embedded inference efficiency on Jetson AGX Orin} (batch size one, MAXN, mixed precision). Latency averages closed-loop inference calls, excluding simulation and communication; memory and FLOPs use fixed-input profiling. DS is from one complete 220-route Bench2Drive evaluation. $^{\dagger}$LM unloaded. Runtime switching uses 1.69 GiB.}
\label{tab:embedded_efficiency}
\footnotesize
\setlength{\tabcolsep}{2.5pt}
\resizebox{\linewidth}{!}{%
\begin{tabular}{@{}lrrrrrr@{}}
\toprule
\textbf{Method} &
\shortstack{\textbf{Params}\\\textbf{(M)}} &
\textbf{DS$\uparrow$} &
\shortstack{\textbf{Latency}\\\textbf{(ms)$\downarrow$}} &
\textbf{FPS$\uparrow$} &
\shortstack{\textbf{GPU mem.}\\\textbf{(GiB)$\downarrow$}} &
\shortstack{\textbf{FLOPs}\\\textbf{(G)$\downarrow$}} \\
\midrule
\textbf{DRIFT w/o LM}
    & \textbf{120}
    & \textbf{96.07}
    & \textbf{35.07}
    & \textbf{28.52}
    & \textbf{0.48}$^{\dagger}$
    & \textbf{150.6} \\
\textbf{DRIFT w/ LM (RR=1)}
    & 255
    & 96.03
    & 172.03
    & 5.81
    & 1.69
    & 308.5 \\
SimLingo~\cite{renz2025simlingo}
    & 1000
    & 86.81
    & 417.81
    & 2.39
    & 2.89
    & 2338.6 \\
\bottomrule
\end{tabular}%
}
\end{table}

\subsection{Real-world applicability}
To show that DRIFT is directly applicable to real-world scenarios, we evaluate it on NAVSIM and AlpaSim. On NAVSIM (\cref{tab:navsim_ltf}), we first compare against recent VLM-based planners that do not directly optimize the task score, following~\cite{nguyen2026lead}. DRIFT without the LM reaches 87.4 PDMS with only 125M parameters. Among the reported VLM-based planners, which are 16--64$\times$ larger (2B--8B), it outperforms three and ties with SGDrive (87.4 PDMS). Compared with the recently proposed Qwen-Drive-1.0~\cite{zhou2026qwen}, a 4B VLM planner trained on 24 public VQA datasets, which we re-evaluated ourselves, DRIFT comes within 0.5 PDMS on navtest (87.4 vs.\ 87.9) at $1/32$ of its size, and is ahead on the harder navhard split (32.5 vs.\ 30.5 EPDMS). Next, we compare against the direct baseline LTFv6~\cite{nguyen2026lead} with the same RegNetY-032 backbone, where DRIFT is on par in PDMS ($+0.4$) and ahead in EPDMS ($+2.3$). Enabling the language model does not degrade performance (87.3 PDMS and 32.8 EPDMS, within noise of the planner alone) while adding explainability. Evaluation details and full results are provided in the supplementary. On the closed-loop AlpaSim, DRIFT scores 89.5 with and without the LM, so detachability carries over to real-world scenes. It is comparable to the LTFv6 baseline (88.6, within the $\pm$1.4 scene-level interval) and below Qwen-Drive (91.7, SFT), which sees more cameras and frames, whereas DRIFT only receives a coarse command. It is also comparable to the very recent TransFuser-K1~\cite{dauner2026123d} (89.6 without navigation), which is trained on significantly more data.

\begin{table}[t]
\centering
\caption{\textbf{Real-world benchmarks.} PDMS (NAVSIM navtest), EPDMS (NAVSIM navhard) and AlpaSim score (nuPlan track). $^\ddagger$: with a test-time heading post-processing (heading set to the tangent of the predicted waypoints). Best in bold, second best underlined, and only methods without direct optimization on NAVSIM score are listed.}
\label{tab:navsim_ltf}
\scriptsize
\setlength{\tabcolsep}{3pt}
\resizebox{\columnwidth}{!}{%
\begin{tabular}{l c c c c}
\toprule
\textbf{Method} & \textbf{Params} & \textbf{PDMS $\uparrow$} & \textbf{EPDMS $\uparrow$} & \textbf{AlpaSim Score $\uparrow$} \\
\midrule
FSDrive~\cite{zeng2026futuresightdrive} & 2B & 85.1 & - & - \\
AdaThinkDrive~\cite{luo2025adathinkdrive} & 8B & 86.2 & - & - \\
FLARE~\cite{xie2026latentvla} & 4B & 86.9 & - & - \\
SGDrive~\cite{li2026sgdrive} & 2B & \underline{87.4} & 27.1 & - \\
Qwen-Drive-1.0, SFT~\cite{zhou2026qwen} & 4B & \textbf{87.9} & 30.5 & \textbf{91.7} \\
\midrule
LTFv6 (reported)~\cite{nguyen2026lead} & 69M & 85.4 & 28.3 & - \\ 
LTFv6 RegNetY-032$^\ddagger$ & 117M & 87.0 & 30.2 & 88.6 \\ 
\textbf{DRIFT w/o LM}$^\ddagger$ & 125M & \underline{87.4} & \underline{32.5} & \underline{89.5} \\
\textbf{DRIFT w/ LM}$^\ddagger$ & 260M & 87.3 & \textbf{32.8} & \underline{89.5} \\
\bottomrule
\end{tabular}%
}
\end{table}

\subsection{Ablation study}

\subsubsection{Feature selection and projectors}
We first study which backbone features the LM should read (\cref{tab:feature_selection}, top). On the LEAD validation split, all variants exhibit strong performances, showing that the \textbf{intermediate driving features already enable strong language decoding}: even LiDAR-only features reach 94.88 SPICE. The largest gain comes from replacing LiDAR with RGB features (METEOR 63.47 $\rightarrow$ 64.38), as RGB carries fine-grained cues (colors, lane types, traffic lights and signs) that are not possible or hard to decode from LiDAR. Combining LiDAR with RGB gives only a negligible improvement, yet we retain LiDAR features since the planner requires it, and sharing the features between the LM and the planner keeps the language grounded in what the planner observes. The ego-speed and target-point projectors give a small gain, concentrated in the speed projector. We keep the target point since it conveys the route intent on which action commentary depends (e.g., which way to turn), although this is not visible in the metrics averaged over all question types.

\begin{table}[t]
  \centering
    \caption{Ablation on feature selection (top) and projector design (bottom). Evaluated after Stage~2 on the in-distribution LEAD validation split, hence not comparable to \cref{tab:drivelm_vqa}. LiDAR refers to $F_{bev}^{L}$ and RGB to $F_{img}^{L}$. $\dagger$ marks the adopted default. B-4 = BLEU-4, S = SPICE, M = METEOR.}
  \label{tab:feature_selection}
  \footnotesize
  \setlength{\tabcolsep}{3pt}
  \resizebox{0.7\linewidth}{!}{%
  \begin{tabular}{@{}llccc@{}}
    \toprule
     & \textbf{Variant} & \textbf{B-4}$\uparrow$ & \textbf{S}$\uparrow$ & \textbf{M}$\uparrow$ \\
    \midrule
    \multirow{5}{*}{\rotatebox{90}{Features}}
      & LiDAR only            & 89.47 & 94.88 & 63.47 \\
      & RGB only              & 90.51 & 95.37 & 64.38 \\
      & RGB + LiDAR           & 90.62 & 95.41 & 64.48 \\
      & \quad + speed         & 90.80 & 95.54 & 64.65 \\
      & \quad + speed, tp$^{\dagger}$ & \textbf{90.87} & \textbf{95.57} & \textbf{64.66} \\
    \cmidrule(lr){1-5}
    \multirow{4}{*}{\rotatebox{90}{Projector}}
      & Cross-attention      & 88.38 & 94.31 & 62.45 \\
      & Deformable attention  & 77.58 & 88.78 & 53.45 \\
      & Spatial attention     & 88.19 & 94.17 & 62.18 \\
      & MLP$^{\dagger}$       & \textbf{90.87} & \textbf{95.57} & \textbf{64.66} \\
    \bottomrule
  \end{tabular}
  }
\end{table}

We next study how the driving features should be projected into the language input space (\cref{tab:feature_selection}, bottom) and compare a two-layer MLP, which keeps all visual tokens, with three attention-based projectors that compress them into learned queries: cross-attention (128 queries), deformable attention (64 queries), and a spatial variant with a 2D locality bias (128 queries). \textbf{The simple MLP outperforms all attention-based projectors on all metrics}, with the deformable variant much worse (METEOR 53.45) and the spatial variant on par with plain cross-attention (62.18 vs.\ 62.45). Hence, the selected features need no additional attention stage or compression before entering the LM, and the gap between the backbone's feature space and the LM's embedding space is small enough for a lightweight per-token mapping to bridge.

\subsubsection{Training stages}
We analyze the impact of each training stage in \cref{tab:recipe_short}. All variants start from the same perception backbone, pre-trained for 30 epochs in Stage~1. DRIFT-Base is the driving-only model without any language, trained for 31 epochs E2E. Adding 2 epochs of VQA pre-training before the same E2E training improves DS from 95.45 to 95.94 (+0.49). This gain is modest given the run-to-run variance, but it appears where expected: in scenarios with vulnerable road users, DS rises from 94.13 to 97.75 and pedestrian collisions drop from 7 to 2. Although the BEV semantic decoders of DRIFT-Base mostly work reliably, \Cref{fig:vru_qual} shows an example of such failures, with more in the supplementary: DRIFT-Base barely detects a crossing pedestrian, whereas after VQA pre-training the pedestrian is clearly grounded. The LM of this variant, however, has been finetuned on driving VQA for only 2 epochs and still answers ``accelerate''. Since this LM is not trained during E2E driving, we do not evaluate its LM path. Our final model therefore adds 16 epochs of joint commentary training, in which the surrogate reasoning is trained to match the LM's. This sharpens the reasoning (``remain stopped due to the pedestrian'') and raises the commentary SPICE from 76.90 to 82.20, at no cost for the surrogate path (95.97 vs.\ 95.94). The resulting LM path drives equally well when invoked at every frame (96.06 DS) and reaches 97.03 DS at RR=2.

\begin{table}[t]
\centering
\caption{\textbf{Effect of training stages.} Epochs of Stage~2 (VQA), 3a (E2E driving) and 3b (driving with commentary training). S: SPICE of the commentary on the LEAD evaluation samples. DS on Bench2Drive; w/ LM uses RR=2.}
\label{tab:recipe_short}
\footnotesize
\setlength{\tabcolsep}{3pt}
\resizebox{\columnwidth}{!}{%
\begin{tabular}{@{}l c c c c c c@{}}
\toprule
 & \multicolumn{3}{c}{Epochs} & & \multicolumn{2}{c}{DS} \\
\cmidrule(lr){2-4}\cmidrule(l){6-7}
Variant & S2 & S3a & S3b & S & w/o LM & w/ LM \\
\midrule
DRIFT-Base & 0 & 31 & 0 & $-$ & $95.45_{\pm 0.26}$ & $-$ \\
+ VQA pre-training & 2 & 31 & 0 & 76.90 & $95.94_{\pm 0.29}$ & $-$ \\
+ Commentary training (\textbf{ours}) & 2 & 15 & 16 & 82.20 & $95.97_{\pm 0.65}$ & $97.03_{\pm 0.09}$ \\
\bottomrule
\end{tabular}%
}
\end{table}

\begin{figure}[t]
  \centering
  \setlength{\tabcolsep}{1pt}
  \newcommand{\vqh}[1]{\makebox[0.31\columnwidth]{\scriptsize #1}}
  \newcommand{\vqt}[1]{\parbox[t]{0.31\columnwidth}{\centering\tiny #1}}
  \begin{tabular}{@{}ccc@{}}
    \multicolumn{2}{c}{\footnotesize Front camera} & \footnotesize GT BEV \\[0pt]
    \multicolumn{2}{c}{\includegraphics[width=0.64\columnwidth]{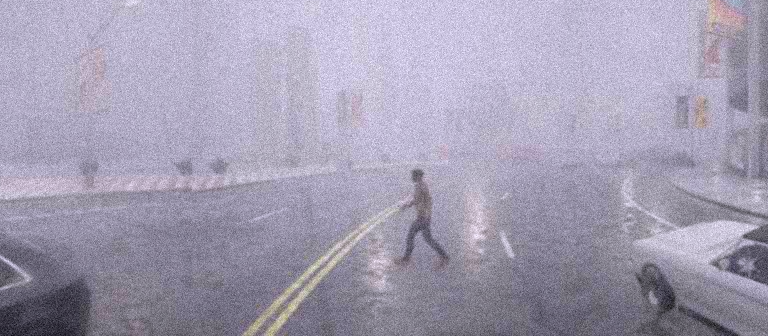}} & \includegraphics[width=0.32\columnwidth]{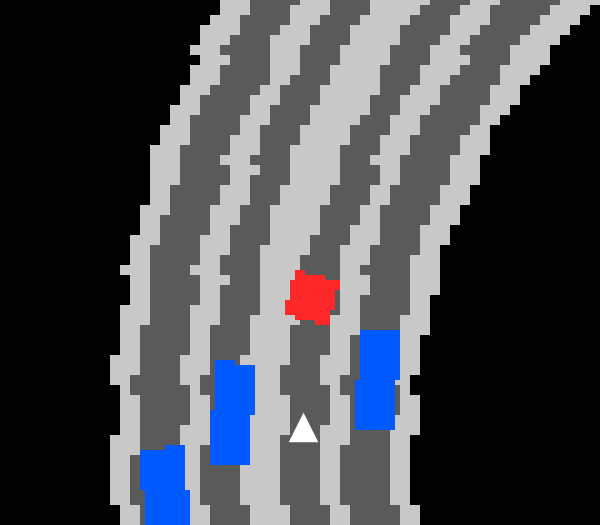} \\[1pt]
    \vqh{DRIFT-Base} & \vqh{+ VQA pre-training} & \vqh{+ Commentary (ours)} \\[0pt]
    \includegraphics[width=0.3\columnwidth]{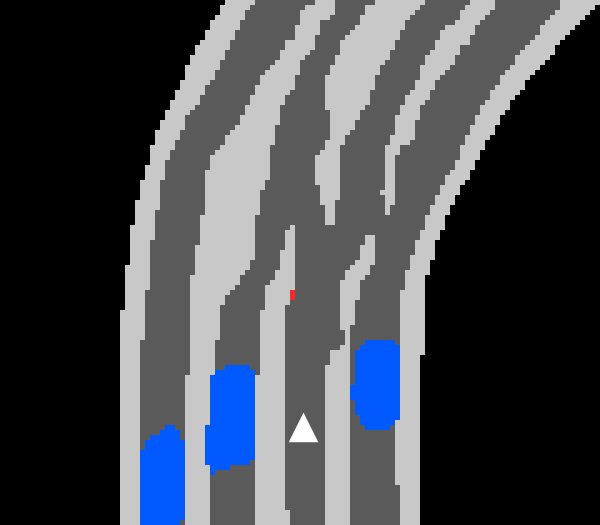} & \includegraphics[width=0.3\columnwidth]{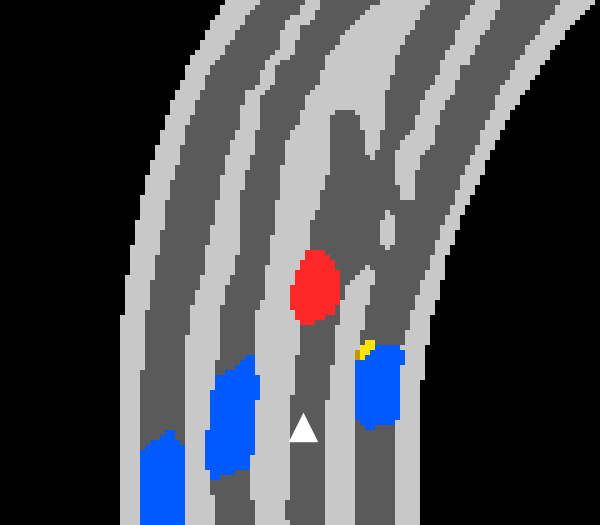} & \includegraphics[width=0.3\columnwidth]{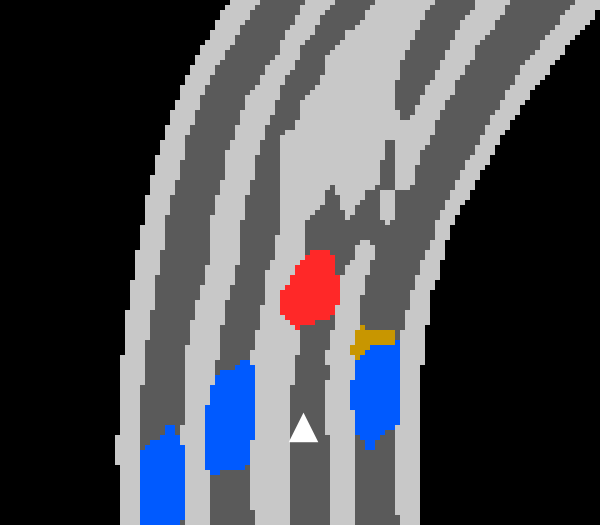} \\[0pt]
    \vqt{\textit{(no language)}} & \vqt{Follow the route. \textbf{\textcolor{red}{Accelerate}} to drive with the target speed.} & \vqt{Follow the route. Remain \textbf{\textcolor[rgb]{0,0.6,0}{stopped}} due to the pedestrian.} \\
  \end{tabular}
  \caption{\textbf{Perception and reasoning in front of a crossing pedestrian.} Top: front camera and GT BEV. Bottom: predicted BEV semantics of each training variant, with the LM's answer to ``What should the ego do next?'' below.}
  \label{fig:vru_qual}
\end{figure}

\section{Conclusion}
In this paper, we argue that VLM-based E2E driving models become more practical by following two design choices: compactness and detachability. To this end, we introduce DRIFT, which shows that rich intermediate driving features enable effective visual question answering and action commentary with a compact LM of under 150M parameters, connected through a simple per-token projector. A surrogate reasoning module trained via knowledge distillation allows the LM to be detached at inference without sacrificing driving performance. On Bench2Drive, DRIFT with LM reasoning reaches up to 97.03 DS and 89.39\% SR with about $4\times$ fewer parameters than the strongest language-based model. Our analysis shows that VQA training can help encode additional features in the driving backbone. It further shows that linking the planner to the LM's reasoning at inference improves the alignment between the generated language and the planned speed, compared to a decoupled variant. Moreover, benchmarking DRIFT on NAVSIM and AlpaSim shows competitive driving quality with a drastically smaller footprint.

{
    \small
    \bibliographystyle{ieeenat_fullname}
    \bibliography{main}
}

\end{document}